\def\year{2020}\relax
\documentclass[letterpaper]{article} 
\usepackage{aaai20}  
\usepackage{times}  
\usepackage{helvet} 
\usepackage{courier}  
\usepackage[hyphens]{url}  
\usepackage{graphicx} 
\usepackage{graphicx}  
\usepackage{amssymb}
\usepackage{booktabs}
\newcommand{\tabincell}[2]{\begin{tabular}{@{}#1@{}}#2\end{tabular}}

\title{Visual Dialogue State Tracking for Question Generation}
\author{\Large \textbf{Wei Pang, Xiaojie Wang\thanks{Corresponding author.}} \\ 
Center for Intelligence Science and Technology, School of Computer Science, \\Beijing University of Posts and Telecommunication\\ 
{pangweitf,xjwang}@bupt.edu.cn 
}
\begin{document}
\maketitle
\begin{abstract}
GuessWhat?! is a visual dialogue task between a guesser and an oracle. The guesser aims to locate an object supposed by the oracle oneself in an image by asking a sequence of Yes/No questions. Asking proper questions with the progress of dialogue is vital for achieving successful final guess. As a result, the progress of dialogue should be properly represented and tracked. Previous models for question generation pay less attention on the representation and tracking of dialogue states, and therefore are prone to asking low quality questions such as repeated questions. This paper proposes visual dialogue state tracking (VDST) based method for question generation. A visual dialogue state is defined as the distribution on objects in the image as well as representations of objects. Representations of objects are updated with the change of the distribution on objects. An object-difference based attention is used to decode new question. The distribution on objects is updated by comparing the question-answer pair and objects. Experimental results on GuessWhat?! dataset show that our model significantly outperforms existing methods and achieves new state-of-the-art performance. It is also noticeable that our model reduces the rate of repeated questions from more than 50\% to 21.9\% compared with previous state-of-the-art methods.
\end{abstract}

\section{Introduction}
\begin{figure}[ht!] \centering
\includegraphics[width=0.95\columnwidth]{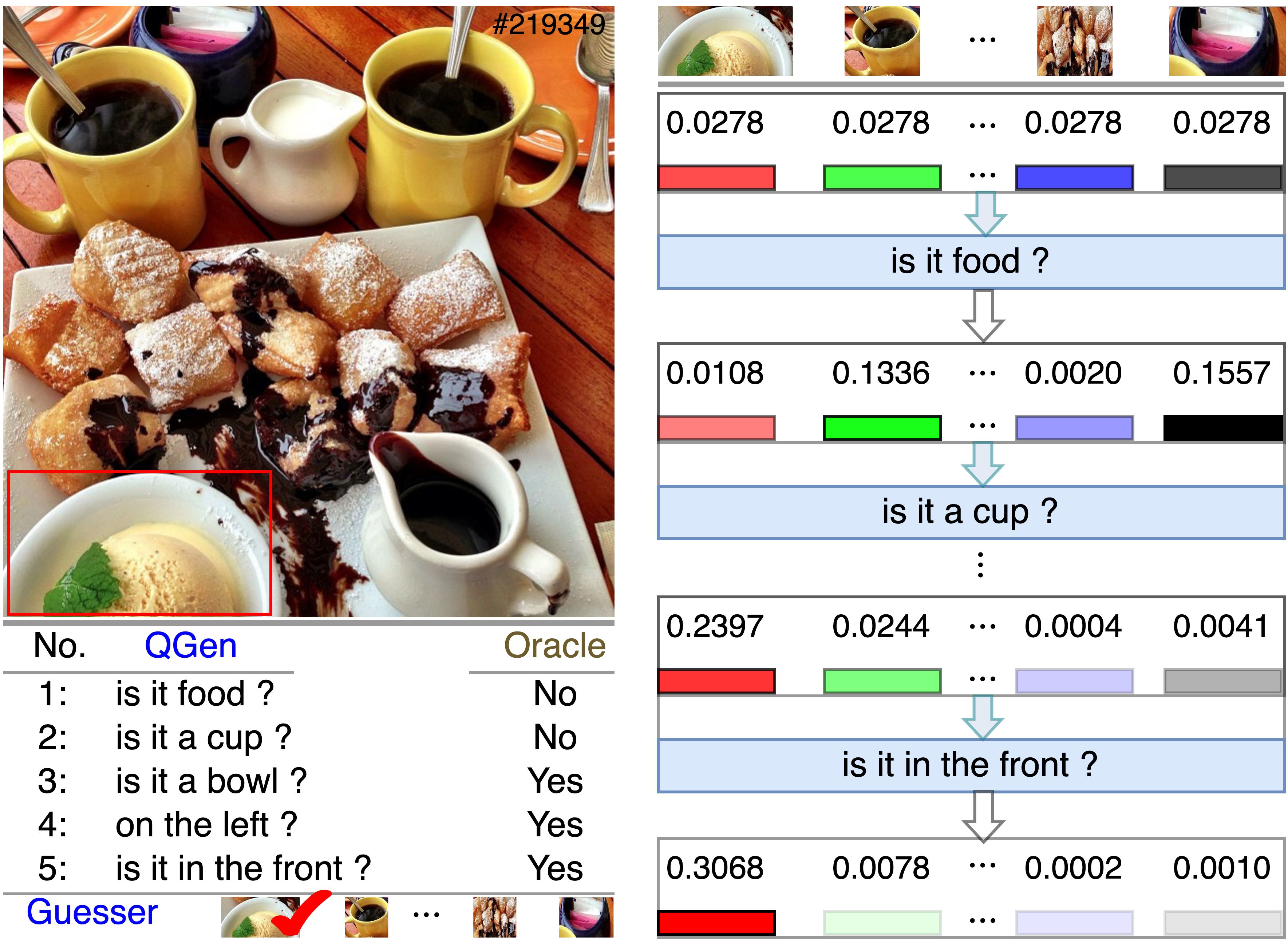}
\caption{The left part gives an example of GuessWhat?! game. The right part illustrates updates of visual dialogue states, including the distribution on objects (numbers on colorful strips) as well as representations of objects (colorful strips).}\label{fig_demonstration}
\end{figure}
Visual dialogue is an increasing interest of multi-modal task, which involves vision, language and reasoning between them in a continuous conversation. Recently, researchers have put forward many different tasks of visual dialogue, including VisDial \cite{CVPR17VisDual}, GuessWhat?! \cite{CVPR17GuessWhat}, GuessWhich \cite{HCOMP17GuessWhich}, Image-Grounded Conversation (IGC) \cite{CVPR17IGC}, Multimodal dialogs (MMD) \cite{AAAI18MMD} and so on. Among them, GuessWhat?! is an object-guessing game played by two players - a Questioner and an Oracle. Given an image including several objects, the goal of the Questioner is to locate a target object - supposed by the oracle oneself at the beginning of a game by asking a series of Yes/No questions. Specially, the Questioner has two sub-tasks, one is Question Generator (QGen) that generates visually grounded questions, the other is Guesser that is to identify the target object given the full dialog context. The Oracle answers questions with yes or no. The left part of Fig.\ref{fig_demonstration} shows a game played by the QGen, Oracle and Guesser. Among them, QGen plays the most important role. Much work has been done on QGen \cite{CVPR17GuessWhat,IJCAI17deepRL,IJCAI18TPG,ECCV18RL3Rewards,NIPS18AQM,CoLING18Stop,ISM18,RIG19,NAACL19Jointly,CVPR19Uncertainty}, this paper also focuses on QGen.

De Vries et al. \cite{CVPR17GuessWhat} proposed the first QGen model. It is an encoder-decoder model, where images are encoded by VGG FC8 features, and questions are decoded by an LSTM (Long Short-Term Memory) \cite{LSTM97}. It is trained in a supervised learning (SL) way. Strub et al. \cite{IJCAI17deepRL} introduced reinforcement learning (RL) to train the QGen model and gained significant improvements. Various of reward functions \cite{IJCAI17deepRL,IJCAI18TPG,ECCV18RL3Rewards,ISM18,CVPR19Uncertainty,RIG19} have been proposed for training QGen model towards producing goal-oriented questions.

However, most of previous work employed unchanged visual representations through a dialogue. For example, De Vries and Strub et al. \cite{CVPR17GuessWhat,IJCAI17deepRL} encoded a raw image into a fixed-size vector using pre-trained VGG network and used the same vector as the image feature for question generation in each round of dialog. Zhao et al. \cite{IJCAI18TPG,ECCV18RL3Rewards,CoLING18Stop,ISM18,NAACL19Jointly,RIG19} employed similar methods on dealing with images. Visual information keeps statically unchanged through the dialogue.

Intuitively, with the progress of a GuessWhat?! dialog, a questioner will focus on different parts of an image. For example, initially, the questioner might pay one's attention to all objects in the image equally. When he/she asks a question “Is it a person?” and receives a positive answer, his/her attention on the image will be changed, more attention might be paid on person objects.

More importantly, with the change of attention on different objects, the representation of the objects might be changed. For example, when a positive answer for the question “Is it a person?” is received, the questioner will pay attention on more detailed information of person objects, such as face, cloth. Therefore, the representation of person objects will be changed to reflect these details. While for non-person objects, their representation might become blurred for saving memory resources.

Borrowing an item from text-based dialogue \cite{WordDST14,TRADE19}, we call the 2-tuple $\left \langle \right .$distribution on objects, representations of objects$\left. \right \rangle$ visual dialogue state (visual state or dialogue state in briefly). As the dialog progress, visual state will be changed, which prompts the questioner to ask new questions (as shown in the right part of Fig.\ref{fig_demonstration}).

This paper proposes a visual dialogue state tracking (VDST) based QGen model to implement above ideas. The model includes a visual-language-visual multi-step reasoning cycle. Representations of objects are firstly updated by current distribution of image objects. A comparison between different objects is employed by a difference-operation-based attention which guided by objects themselves. Results of the attention is then used to decode new question. Finally, a cross-modal matching between question-answer pairs and object representations is designed for updating distribution of objects. Experimental results show our model achieves the new state-of-the-art performance on GuessWhat?! task. Especially, our model reduces the rate of repeated questions from more than 50\% to 21.9\% compared with previous state-of-the-art methods.

To summarize, our contributions are mainly three-fold:
\begin{itemize}
\item We propose a visual dialogue state tracking (VDST) based model for question generator in GuessWhat?!. The VDST, including not only distribution of objects but also representations of objects, is tracked and updated with the progress of a dialogue.
\item We achieve new state-of-the-art performances on QGen task of GuessWhat?! on four different training methods including a separate and a joint Supervised Learning, a separate Reinforcement Learning and a Cooperative Learning.
\item We reduce the rate of repeated questions in dialogues from more than 50\% to 21.9\% compared with previous state-of-the-art methods.
\end{itemize}

\section{Related Work}
GuessWhat?! Game is one of the visual dialogue tasks introduced in \cite{CVPR17GuessWhat}, which consists of three sub-tasks: Oracle, QGen and Guesser. Most of work on GuessWhat?! focuses on QGen because it plays a key role in the game.

De Vries et al. \cite{CVPR17GuessWhat} proposed the first QGen model. The entire image is encoded as a fixed-size vector (e.g., 1000-dim) that is extracted from VGG-fc8 layer \cite{VGG15}. The previous dialogues, such as question-answer pairs, are encoded by a hierarchical recurrent encoder. Question generation is performed by an LSTM where input is a concatenation of above VGG feature and representation of previous dialogues. They cast QGen task as a supervised learning (SL) task, and the model is trained by minimizing cross-entropy loss between the predicted question and the ground-truth question.

There are mainly two lines of ongoing work for QGen task. One is on model learning. Strub et al. \cite{IJCAI17deepRL} first cast QGen as a reinforcement learning (RL) task, they use a zero-one reward to lead QGen to produce goal-oriented questions, and achieve a significant improvement over the supervised model \cite{CVPR17GuessWhat}. Zhang et al. \cite{ECCV18RL3Rewards} add two intermediate rewards. Recently, to better explore the space of actions (word selections), Abbasnejad et al. \cite{CVPR19Uncertainty} design reward for each action based on a Bayesian model. Zhao et al. \cite{IJCAI18TPG} calculate a temperature for each action based on action frequencies. In addition, Zhao et al. first train QGen and Guesser by reinforcement learning from self-play. Shekhar et al. \cite{NAACL19Jointly} cast QGen and Guesser as a multi-task problem, and train the two models by cooperative learning (CL) \cite{ICCV17VisDi}.

Lee et al. \cite{NIPS18AQM} use information gain to sample questions on the GuessWhat?! dataset, rather than to generate new questiones. Shukla et al. \cite{RIG19} use regularized information gain to reward their QGen model.

The other line of work for QGen task is on encoding and fusion of visual and linguistic information through a dialogue. Following \cite{CVPR17GuessWhat}, the methods in \cite{IJCAI17deepRL,Transferring17,IJCAI18TPG,ECCV18RL3Rewards,NIPS18AQM,CoLING18Stop,WordDST14,RIG19,CVPR19Uncertainty} all obtain the image features from VGG fc-8 layer. Shekhar et al. \cite{NAACL19Jointly} develop a common grounded dialogue state encoder that fuse both image features and linguistic representations, where the image features (e.g., 2048-dim) are extracted from the second to last layer of ResNet152 \cite{ResNet16}, the linguistic representation is obtained by an LSTM which takes input the previous dialogue history. The two types of information are concatenated and passed through a multi-layer perceptron (MLP) layer to get an embedding, which is called the dialogue state and given as input to both QGen and Guesser.

Whatever the encoder used in above models, for a given image, the encoded vector for the image keeps fixed through a dialogue. Different from previous work, the model proposed in the paper will have changed representation of objects in each round of dialog. We will show it will bring significant improvement to question generation.

\section{Model: Visual Dialogue State Tracking}
\begin{figure*}[htb!] \centering
\includegraphics[width=2.0\columnwidth]{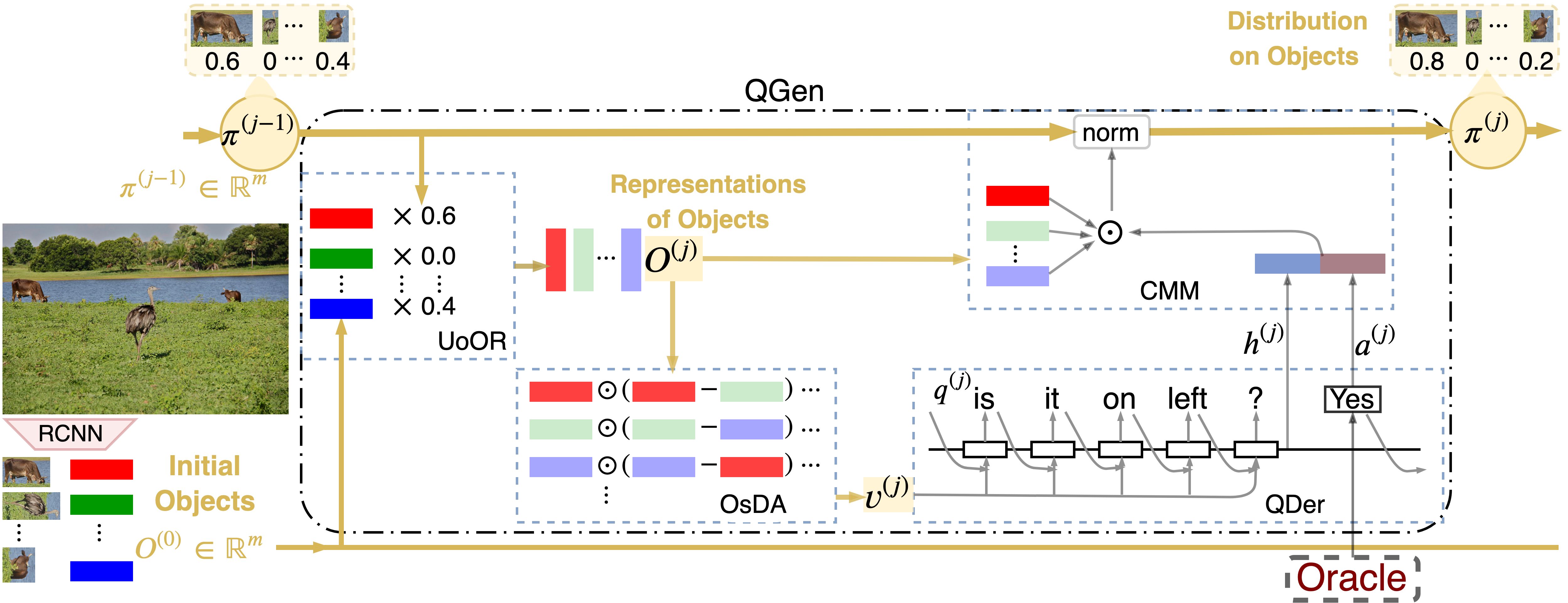}
\caption{Overall structure of VDST model.}\label{fig_structure}
\end{figure*}

The framework of our visual dialogue state tracking model (for short, VDST) is illustrated in Fig.\ref{fig_structure}. In each round of dialog, there are four modules for QGen: Update of Object Representation (UoOR), Object-self Difference Attention (OsDA), Question Decoder (QDer) and Cross-Modal Matching (CMM). Where, UoOR updates representations of objects according to the distribution of objects, new representations are used by OsDA to induce attention vectors which are then used to generate new question by QDer, finally, CMM is employed to update the distribution of objects. Details of each module are given respectively in remainders of this sections.

\subsection{Update of Object Representation (UoOR)}
Object representations for bounding boxes or image regions are extracted via Faster-RCNN \cite{RCNN} as in Eq.\ref{eqn_rcnn}.
\begin{equation} \label{eqn_rcnn}
\centering
O^{f} = \mathrm{RCNN}(image),
\end{equation}
\noindent where $O^{f} \in \mathbb{R}^{m\times (d_{s} + 8)}$ includes $m$ objects $\{o_{k}\}_{k=1}^{m}$. For each object $o_{k}$, its representation is a concatenation of $d_{s}$-dim static feature provided by bottom-up-attention \cite{BottomUp} and 8-dim vector of the spatial coordinates $[x_{min}$, $y_{min}$, $x_{max}$, $y_{max}$, $x_{center}$, $y_{center}$, $w_{box}$, $h_{box}]$, where $x$, $y$ denote the coordinates and $w_{box}$, $h_{box}$ denote the width and height of an object. Following \cite{CVPR17GuessWhat}, the coordinates are normalized by the image width and height so that its values range from -1 to 1.

To map $o_{k}$ and embedding of words to the same dimension $d$, an MLP layer is used as in Eq.\ref{eq_image},
\begin{equation} \label{eq_image}
\centering
O^{(0)} = \mathrm{MLP}(O^{f}),
\end{equation}
where $O^{(0)}\in \mathbb{R}^{m\times d}$ is viewed as a set of initial object representations for objects.

Let $\pi^{(j)}\in \mathbb{R}^{m}$ be probability distribution (or attention distribution) on m objects at the jth dialogue round. It gives the accumulative attention paid on each object up to jth round dialogue. Please note that $\pi^{(j)}$ is the attention distribution instead of a distribution of each object as target object which will be calculated by the Guesser. Let the initial distribution $\pi^{(0)}$ be a uniform distribution, which means each object has been paid equal attention at the beginning. Update of Object Representation (UoOR) is defined as in Eq.\ref{eqn_bn}.

\begin{equation} \label{eqn_bn}
\centering
O^{(j)} = ({\pi^{(j)}})^{T} O^{(0)},
\end{equation}
\noindent where $O^{(j)}\in \mathbb{R}^{m\times d}$ denotes the representations of objects at the jth round.

From Eq.\ref{eqn_bn}, we can observe that representations of objects are updated by the distribution of objects at each dialog round. Because the distribution of objects is updated at each dialog round (will be specified in CMM module), the representation for a same object at different dialog round is therefore different. It models the intuition: with the change of attention on different objects, the representation of the objects might be changed.

\subsection{Object-self Difference Attention (OsDA)}

We define Object-self Difference Attention (OsDA) at jth round of dialogue as in Eq.\ref{eqn_OsDA},
\begin{equation}  \label{eqn_OsDA}
\centering
v^{(j)} = \mathrm{softmax}([o^{(j)}_{i}\odot (o^{(j)}_{i} - o^{(j)}_{k})]W)^{T}O^{(j)}.
\end{equation}
\noindent Where $o^{(j)}_{i}\in O^{(j)}$, $o^{(j)}_{i}\odot (o^{(j)}_{i} - o^{(j)}_{k})\in \mathbb{R}^{d}$ is difference between ith object and kth object under the guidance of the ith object itself, $W\in \mathbb{R}^{md\times g}$ is a parameter matrix to be learned with $g$ glimpses. The differences between the $i$th object and all other objects are concatenated into a $md$-dimensional vector. For m objects,  operation $[o^{(j)}_{i}\odot (o^{(j)}_{i} - o^{(j)}_{k})]\in \mathbb{R}^{m\times md}$ constructs a $m\times md$ matrix. The softmax transforms the glimpse results of differences to attention weight, and then output the final attentive visual feature $v^{(j)}$. It will be used as visual context to decode a question for next round of dialog.

We have two notes on OsDA.

Firstly, visual comparison between objects could provide visually discriminative cues and help quickly distinguish multiple objects in an image. Motivated by that, for a specify object $o^{(j)}_{i}$, comparison to the kth object as the subtraction of $o^{(j)}_{i}$ and $o^{(j)}_{k}$ under guidance of itself, namely the difference operation $o^{(j)}_{i}\odot(o^{(j)}_{i}-o^{(j)}_{k})$, is designed.

Secondly, we aggregate their differences and map them with g glimpse to an attention distribution over objects, and then we get the final visual context $v^{(j)}$. Note that here $v^{(j)}$ is dynamically changed at different round with the UoOR.

\subsection{Question Decoder (QDer)}
A one-layer LSTM is employed as QDer for generating questions, as shown in Eq.\ref{eqn_LSTM},
\begin{equation} \label{eqn_LSTM}
\centering
w^{(j)}_{i+1} = \mathrm{LSTM}([v^{(j)}; w^{(j)}_{i}]),
\end{equation}
\noindent where $w^{(j)}_{i}$ denotes the ith word of question $q^{(j)}$, $[;]$ refers to concatenation. Note that the last hidden state $h^{(j)}$ of the LSTM decoder is used as the representation of $q^{(j)}$ as shown in Fig.\ref{fig_structure}.

\subsection{Cross-Modal Matching (CMM)}
When answer $a^{(j)}$ for question $q^{(j)}$ is received from the Oracle, Questioner needs to renew his/her judgement on which object should be paid more attention, i.e. the probability distribution on m objects, according to new information including in the question-answer pair. Three steps are designed for the update.

Firstly, the embedding of answer $a^{(j)}$ is concatenated to the representation of question, $h^{(j)}$, to form a representation of question-answer pair: $h^{(j)}_{a}=[h^{(j)}; a^{(j)}]$ as shown in Fig.\ref{fig_structure}. 

Secondly, a Cross-Modal Matching (CMM) network is defined as in Eq.\ref{eqn_cmm} to measure how many changes the new question-answer pair $\left \langle q^{(j)}, a^{(j)} \right \rangle$ bring to the distribution on objects.

\begin{equation} \label{eqn_cmm} \centering
\hat{\pi}^{(j)} = \mathrm{softmax}(\frac{\tanh(O^{(j)}U^{T}\odot V^{T}h^{(j)}_{a})}{\sqrt{d}})
\end{equation}
\noindent Where $U\in \mathbb{R}^{d\times d}$, $V\in \mathbb{R}^{2d\times d}$ are learnable parameters, $\sqrt{d}$ in denominator is used to control variance. $\odot$ denotes Hadamard product used for fusion of multimodal information from visual object representations and textual question-answer pair. A softmax function is used to produce a probability distribution $\hat{\pi}^{(j)}$ over m objects, which is the change that $\left \langle q^{(j)}, a^{(j)} \right \rangle$ brings.

Thirdly, given the change $\hat{\pi}^{(j)}$ in this round and the accumulative probability distribution $\pi^{(j-1)}$ on objects at previous round. $\pi^{(j)}$, the accumulative probability distribution at jth round is defined as in Eq.\ref{eqn_norm},
\begin{equation} \label{eqn_norm}
\centering
\pi^{(j)} = \mathrm{norm}(\pi^{(j-1)}\cdot \hat{\pi}^{(j)}),
\end{equation}
where $\mathrm{norm}$ is normalization operation to keep $\pi^{(j)}$ be a probability distribution.

\section{Model Training}
Because the paper focuses on visual state tracking, we following existing way for model training. As in \cite{IJCAI17deepRL}, we first train QGen, Oracle and Guesser independently in supervised way with a cross-entropy loss, and then tune the QGen by RL with the oracle and guesser fixed. 

The  cross-entropy loss is defined as in Eq.\ref{equ_SLloss},
\begin{equation} \label{equ_SLloss}
\centering
L_{SL}(\theta) = \sum^{T}_{t=1}-\log(p(w_{t})),
\end{equation}
\noindent where T is the length of a complete dialogue, $p(w_{t})$ is the probability of the ground-truth word at step t in the dialogue, $\theta$ denotes the parameters.

A zero-one reward is used for RL. i.e. $r_{D} = 1$ for a successful guessing, otherwise, $r_{D} = 0$. The RL loss is defined as the negative expected reward:
\begin{equation} \label{equ_RLloss}
\centering
L_{RL}(\theta) = -\mathbb{E}_{\pi_{\theta}}[\sum^{T}_{t=1}(r_{A}(a_{t}) - b)\log\pi_{\theta}(a_{t})],
\end{equation}
\noindent where b is the baseline function the same as \cite{IJCAI17deepRL}, $a_{t}$ denotes an action (i.e. word) choice at each time step $t$. In the case of reward $r_{A}(a_{t})$ for each action, we spread the reward $r_{D}$ uniformly over the sequence of actions. The model can be trained end-to-end by minimizing the loss in Eq.\ref{equ_RLloss}.

\section{Experiments and Analysis}

\noindent {\bf Dataset and Evaluation}	We evaluate our model on the GuessWhat?! dataset introduced in \cite{CVPR17GuessWhat}. This dataset has 155k dialogues on 66k images and consists of 822k question/answer pairs, the vocabulary contains nearly 5k words. We use the standard partition of the dataset to the training (70\%), validation (15\%) and test (15\%) set as in \cite{CVPR17GuessWhat,IJCAI17deepRL}. 

Following \cite{IJCAI17deepRL}, we report the task success rate by three inference methods, sampling, greedy and beam search (beam size 20) on New Object and New Game respectively. New Object refers to the target object is new but the image has been seen in training, while New Game refers to the image and target are all previously unseen.

\noindent {\bf Implementation Details}	We limit the maximum number of questions to 5 (5Q) \cite{CVPR17GuessWhat} and 8 (8Q) \cite{IJCAI17deepRL,IJCAI18TPG,NAACL19Jointly,RIG19} respectively for comparisons. We set the maximum number of words to 12 as \cite{IJCAI17deepRL}. For each image, 36 bounding boxes (objects) are extracted from RCNN, image features are $36\times 2056$. The dimension of word embedding is 512. The image features are mapped to 36x512 with a fully-connection layer follows a swish activation \cite{Swish17}. The LSTM hidden unit number in QDer is 512. Attention glimpse is 2. Our code and other materials will be published in the near future.

{\bf Supervised Learning}	We train the Guesser and Oracle model for 30 epochs, and pre-train the QGen model for 50 epochs, using Adam optimizer \cite{Adam15} with a learning rate of 1e-4 and a batch size of 64. Note that pre-training the QGen is important to quickly approximate a relatively good policy for RL fine-tuning.

{\bf Reinforcement Learning}	We follow the same RL agent environment as \cite{IJCAI17deepRL}. After supervised training, we then initialize the QGen environment with the pre-trained parameters from the above supervised learning, and post-train the QGen model with REINFORCE \cite{reinforce92,PG2000} for 500 epochs, using stochastic gradient descent (SGD) with a learning rate of 1e-3 and a batch size of 64.

{\bf Baseline and SOTA models}	Four SL based models are compared in experiments, there are the first SL model \cite{CVPR17GuessWhat}, GDSE model \cite{CoLING18Stop,NAACL19Jointly}, the first RL model \cite{IJCAI17deepRL} pre-trained in SL phase and TPG model \cite{IJCAI18TPG} pre-trained in SL phase. Six RL based models are compared, there are the first RL model, TPG model, VQG \cite{ECCV18RL3Rewards}, ISM \cite{ISM18}, Bayesian model \cite{CVPR19Uncertainty} and RIG model \cite{RIG19}. Where the RL, TPG and RIG model used a zero-one reward, which is as the same as that in our model. The VQG model used more considerate rewards which integrates three intermediate rewards for policy learning, including goal-achieved reward (i.e. zero-one reward), progressive reward and informativeness reward.

\begin{table*}[htb]
\centering
\caption{Task success rate for each model.} \label{tab_result1}
\begin{tabular}{l|c|c|c|c}
\hline
\multicolumn{5}{c}{New Object} \\ \cline{1-5}
& Sampling & Greedy & BSearch & Max Q's\\
\hline
SL \cite{CVPR17GuessWhat}&41.6&43.5&47.1&5\\
\cite{IJCAI17deepRL} (pre-trained in SL) & - & 44.6 & - & 8\\
\cite{IJCAI18TPG} (pre-trained in SL)&-&48.77&-&8\\
VDST(ours, pre-trained in SL)&45.02&\textbf{49.49}&-&5\\
VDST(ours, pre-trained in SL)&\textbf{46.70}&48.01&-&8\\
\hline
RL \cite{IJCAI17deepRL} & 58.5 & 60.3 & 60.2 & 5\\
RL \cite{IJCAI17deepRL} & 62.8 & 58.2 & 53.9 & 8\\
TPG (MN)\cite{IJCAI18TPG}   & 62.6 & -  & - & 5\\
VQG \cite{ECCV18RL3Rewards} & 63.2 & 63.6 & 63.9 & 5\\
ISM \cite{ISM18}&-&64.2&-&-\\
Bayesian \cite{CVPR19Uncertainty} & 61.4 & 62.1 & 63.6 & 5\\
RIG  as rewards \cite{RIG19}&65.20&63.00&63.08&8\\
RIG  as a loss with 0-1 rewards \cite{RIG19}&67.19&63.19&62.57&8\\
VDST(ours) & \textbf{66.22}& \textbf{67.07} &\textbf{67.81}& 5\\
VDST(ours) & \textbf{69.51}& \textbf{70.55} &\textbf{71.03}& 8\\
\hline \hline
\multicolumn{5}{c}{New Game} \\ \cline{1-5}
SL \cite{CVPR17GuessWhat}&39.2&40.8&44.6&5\\
SL \cite{CVPR17GuessWhat}&-&40.7&-&8\\
VDST(ours, pre-trained in SL)&42.92&\textbf{45.94}&-&5\\
VDST(ours, pre-trained in SL)&\textbf{44.24}&45.03&-&8\\
\hline
RL \cite{IJCAI17deepRL} &56.5& 58.4 & 58.4& 5\\
RL \cite{IJCAI17deepRL} & 60.8& 56.3 & 52.0& 8\\
VQG \cite{ECCV18RL3Rewards} & 59.8 & 60.7 & 60.8 & 5\\
ISM \cite{ISM18}&-&62.1&-&-\\
GDSE-SL \cite{NAACL19Jointly} &-& 47.8  &-& 5\\
GDSE-SL \cite{NAACL19Jointly} &-& 49.7  &-& 8\\
GDSE-CL \cite{NAACL19Jointly} &-& 53.7  &-& 5\\
GDSE-CL \cite{NAACL19Jointly} &-& 58.4  &-& 8\\
Bayesian \cite{CVPR19Uncertainty} & 59.0 & 59.8 & 60.6 & 5\\
RIG  as rewards \cite{RIG19}&64.06&59.0&60.21&8\\
RIG  as a loss with 0-1 rewards \cite{RIG19}&65.79&61.18&59.79&8\\
VDST(ours) & \textbf{63.85}& \textbf{64.36} & \textbf{64.44}& 5\\
VDST(ours) & \textbf{66.76}& \textbf{67.73} &\textbf{67.52}& 8\\
\hline
\end{tabular}
\end{table*}

\subsection{Comparisons on Success Rate of Guessing}
We compare our model with the previous state-of-the-art (SOTA) methods on success rate of guessing in the subsection. Tab.1 reports the performance for all models.

The first part of Tab.\ref{tab_result1} shows results by SL. Our model (VDST) outperforms three previous models. The second part of Tab.\ref{tab_result1} shows results by RL. It can be seen that our model not only outperforms all the models that use zero-one reward but also outperforms the models with more considerate reward significantly. Our model achieves a new state-of-the-art results on both New Object and New Game. Specifically, compared with the VQG, Bayesian and RIG models, our model achieves nearly 7 points promotions on New objects on Greedy case, and achieves nearly 6 points promotions on New games on Greedy way. Compared with GDSE-CL, VDST achieves 10.66\% and 9.33\% higher accuracy for 5Q and 8Q. On Greedy case in New games, compared with the RL baseline, VDST improves task success rate by 5.96\% for 5Q and 11.43\% for 8Q; compared with RIG models, our model improves by more than 6.55\%.

Some models employed cooperation learning to co-train QGen and Guesser to promote their successful rate, and achieve even better performance \cite{IJCAI18TPG,CVPR19Uncertainty}. Such as Abbasnejad et al. \cite{CVPR19Uncertainty} post-trained QGen and Guesser with RL and improved the accuracy from 62.1\% to 69.2\% on New Object, which gives a 7.1\% improvement. The paper focuses on QGen itself, and does not implement cooperation learning experiments due to time and space limitation.

In summary, our model consistently achieves the best performances on success rate of guessing on both New objects and New games compares with existing baseline and SOTA models.

\subsection{Comparisons on Quality of the Dialogues}

\begin{table}[htb!]
\centering
\caption{Comparisons on quality of the dialogues.} \label{tab_result3}
\begin{tabular}{lcc}
\hline
              &\tabincell{c}{\%Games\\with\\repeated Q’s} & \tabincell{c}{Lexical\\diversity} \\ \hline
\tabincell{l}{SL\\ \cite{CVPR17GuessWhat}}&93.50&0.030\\ \hline
\tabincell{l}{RL\\ \cite{IJCAI17deepRL}}&96.47&0.073\\ \hline
\tabincell{l}{GDSE-SL\\ \cite{NAACL19Jointly}}&55.80&0.101\\ \hline
\tabincell{l}{GDSE-CL\\ \cite{NAACL19Jointly}}&52.19&0.115\\ \hline
VDST(ours) &\textbf{21.9}&\textbf{0.150}\\ \hline
Human &-&0.731\\
\hline
\end{tabular}
\end{table}

We then compare the quality of dialogues, especially the quality of questions generated by different models with the 8Q setting. We follow \cite{NAACL19Jointly} in which two linguistic measurements: percentages of repeated questions in a dialogue and lexical diversity, are used. Where the former measured as the amount of games that contain at least one repeated question over all games, the latter measured as the percentage of unique words over all games.

{\bf Repeated questions}
Tab.\ref{tab_result3} shows that our model has much lower percentage of repetitive questions in a dialogue than all the other models. Only 21.9\% of dialogues contain repetitive questions, compared to 52.19\% in GDSE-CL and higher ones in other models. This indicate that tracking and update of visual dialogue state in a dialog could improve the quality of the dialog significantly. It enables the guesser shift its attentions to new objects or new features so that it tends to ask new questions.

{\bf Lexical diversity}
Our model also improves lexical diversity from 0.115 in GDSE-CL to 0.15. It doubles the score that in other baseline models. It is only natural that a model asking more new questions on new objects or new features will bring more lexical diversity.

Overall, our model generates significantly fewer repeated questions than all the compared methods, and is therefore with richer words than that in others.

\subsection{Strategy for questions asking learnt by our model}

We further investigate the policy of question asking in dialogues learned by our model. The class labels and a classifier in \cite{NAACL19Jointly} are used. There are entity question (e.g., “is it a person?”) and attribute question (e.g., “is it a red?”). Entity question labels can be either an object category (e.g., car) or a super-category (e.g., vehicle). Attribute question has 6 fine-grained subclasses including color, location, size, shape, texture and action.

\begin{figure}[htb!]
\centering
  \includegraphics[width=0.95\columnwidth]{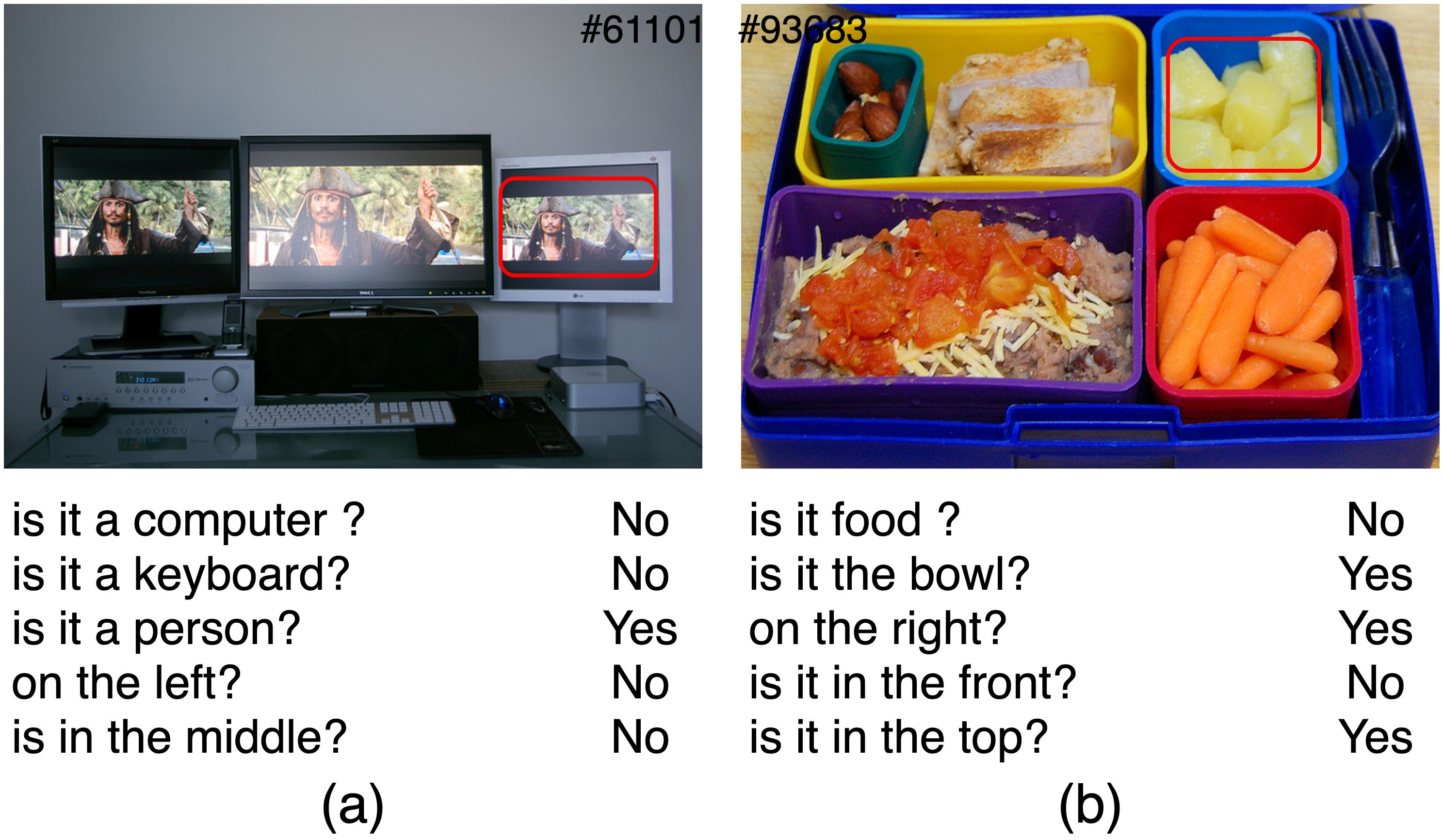}
  \caption{Questions generated by our VDST model. The target object is highlighted in red box.} \label{fig_policy}
\end{figure}

\begin{table}[htb!]
\centering
\caption{Experimental results of ablation studies. Row 2-3 shows the effect of each module by removing it from the full model.} \label{tab_ablation}
\begin{tabular}{llll}
\hline
 & & \multicolumn{2}{c}{New Object} \\  \cmidrule(r){3-4}
\#&Model& Sampling & Greedy \\ \hline
1&VSDT(full model)&66.22& 67.07\\
2&$\quad-$UoOR\&CMM&62.73&64.30\\
3&$\quad-$OsDA&64.06&65.21\\
\cline{3-4}
\cline{3-4}
& &  \multicolumn{2}{c}{New Game} \\ \cline{3-4}
1&VSDT(full model)& 63.85& 64.36\\
2&$\quad-$UoOR\&CMM&60.07&60.93\\
3&$\quad-$OsDA&60.89&61.71\\
\hline
\end{tabular}
\end{table}

When questions in GuessWhat?! human dialogues are labeled with the class labels by the classifier, we find that 22.38\% of sequences of human questions in a single dialogue implement an interesting question asking policy. It begins with an entity question. If an answer “No” for the question is received, it keeps asking an entity question on other categories until an answer “Yes” is received. It then shifts to ask an attribute question such as location questions, and keeps asking this kind of questions to the end of dialog. 

\begin{figure}[h!]
\centering
  \includegraphics[width=0.95\columnwidth]{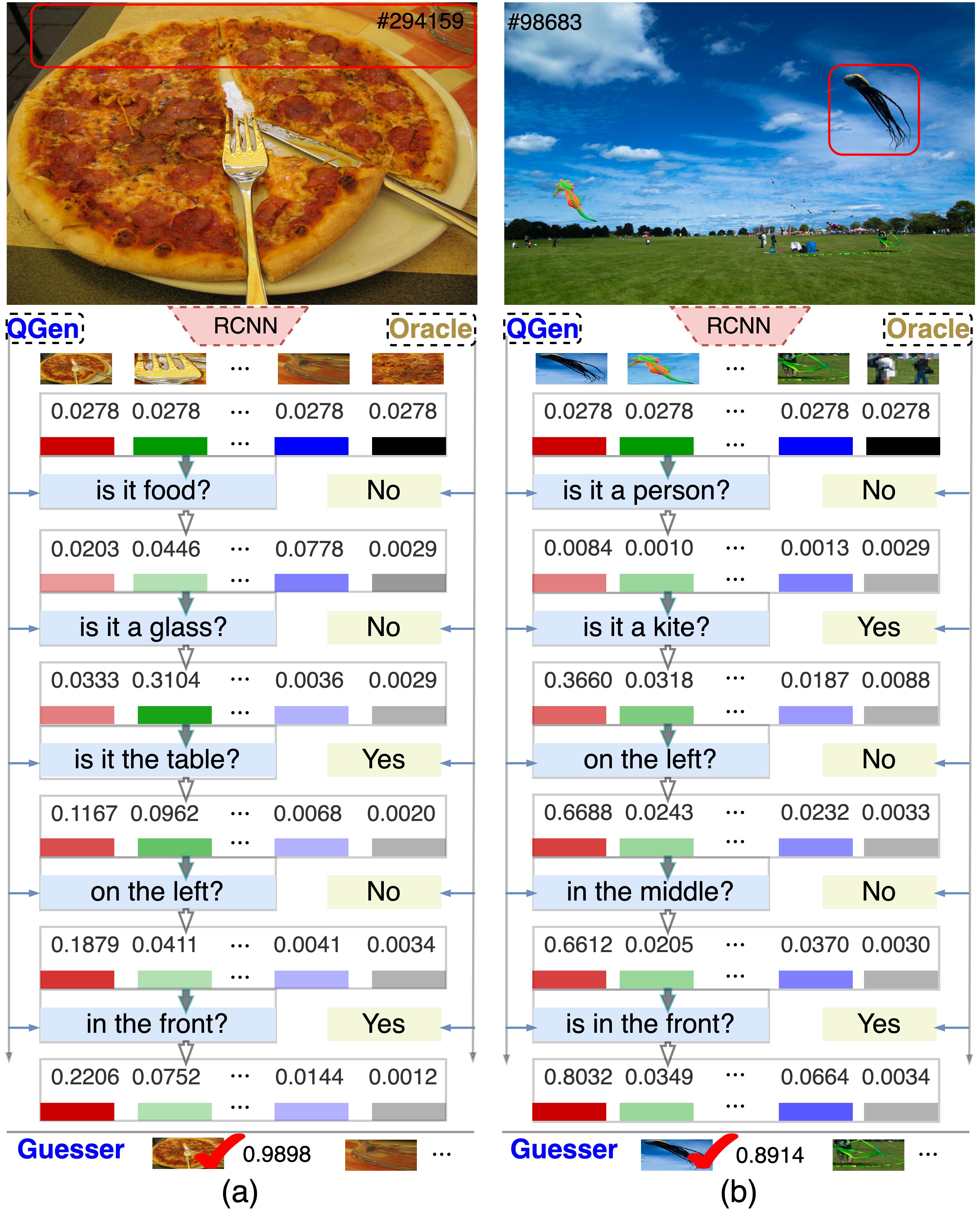}
 \caption{Two successful examples show the tracking procedure of VDST.} \label{fig_tracking}
\end{figure}

We label the questions generated by our model in the same way as in above. We find our model learn the above policy well. In 67.88\% of dialogues, the sequences of questions generated by our model implement this strategy. 

Fig.\ref{fig_policy} gives two successful sequences of questions generated by our model in 5Q setting. In \ref{fig_policy}(a), the agent starts by an entity question “is it a computer?”, and keeps on asking entity question when a “yes” is received. The agent now identifies the category of target object, and then turns to ask attribute questions. Here the location question is chosen. It finally guesses the target object. In \ref{fig_policy}(b), one more success case is shown.

\subsection{Ablation Studies on Major Modules}
In this section, we conduct some ablation experiments to illustrate the effect of each major module in our model. Tab.\ref{tab_ablation} shows the results for 5Q.

Firstly, we considerate the effect of update of the visual dialogue state which consist of the distribution of object and the representations of objects. The distribution of object is updated by CMM module and the representations of objects are updated by UoOR. As we can see, the performance of a model without UoOR and CMM drop nearly 3 points as shown in Tab.\ref{tab_ablation}. It shows visual dialogue state tracking plays a key role in the model. Please note that there is no meaning to employ UoOR without CMM or vice versa, because the visual state in our model is a 2-tuples.

Secondly, we considerate the effect of OsDA which is used to obtain attention for question generation. For removing OsDA, we replace the Eq.\ref{eqn_OsDA} by an average representation of  $O^{(j)}$. Experimental results show OsDA helps the model very much, especially in New Game case.

\subsection{Qualitative Evaluation}
In Fig.\ref{fig_tracking}, we give two examples to visualize the tracking procedure in our model;. We only show 4 objects and other objects are omitted for simplicity in each example. The colorful strips denote representations of objects, and the number values on strips denote their probabilities in the attention distribution on objects.

Taking Fig.\ref{fig_tracking}(a) as an example, we give a simple explanation. At the beginning of the game, representations of all objects are given by RCNN, all objects have equal probability (0.0278 as shown in the figure) as target. Basing on the initial information, the QGen first generates a question “is it food?”, and receives an answer “no” from the oracle. Then the attention probabilities of each object is changed, such as attention on the food object (pizza shown in the last column) drops to 0.0029. After round 3, attention on the target object (shown in the first column) increases to 0.1167. Meanwhile, with its increased attention, the representation of the target object is updated as shown in colour variations. In \ref{fig_tracking}(b), at round 2, the target object has already obtained the maximum attention (36.6\%) compared to other objects, begins at round 3, the agent focuses on asking location questions about the target object, and pays much less attention to others.

\section{Conclusion}

This paper presents a visual dialogue state tracking (VDST) based QGen model. The visual dialogue state, include distribution on objects as well as representations of objects. They are tracked and updated with the progress of a dialogue. With the update of object distribution, the representations of objects are updated, and prompt the questioner to ask new questions. Experimental results on the GuessWhat?! dataset show the model achieves the new state-of-the-art in both settings of supervised learning and reinforcement learning. The model reduces the rate of repeated questions from more than 50\% to 21.9\% compared with previous state-of-the-art methods. There are lots of problems to be resolved in the future, such as how to learn more flexible policy of asking questions, when to stop the question asking and turn to guess.


\section{Acknowledgements}
We thank the reviewers for their comments and suggestions. This paper is supported by NSFC (No. 61906018), Huawei Noah's Ark Lab and MoE-CMCC “Artificial Intelligence” Project (No. MCM20190701).

\bibliographystyle{aaai} \bibliography{GuessWhat20.bib}
\end{document}